\documentclass[sigconf, nonacm]{acmart}

\usepackage{multirow}
\usepackage{booktabs}
\usepackage{adjustbox}
\usepackage{enumitem}
\usepackage{makecell}
\usepackage{subcaption}
\usepackage{balance}
\usepackage[a-1b]{pdfx}

\newcommand\vldbdoi{XX.XX/XXX.XX}
\newcommand\vldbpages{XXX-XXX}
\newcommand\vldbvolume{14}
\newcommand\vldbissue{1}
\newcommand\vldbyear{2020}
\newcommand\vldbauthors{\authors}
\newcommand\vldbtitle{\shorttitle} 
\newcommand\vldbavailabilityurl{URL_TO_YOUR_ARTIFACTS}
\newcommand\vldbpagestyle{plain} 

\begin{document}
\title{HGTUL: A Hypergraph-based Model For Trajectory User Linking }

%%
%% The "author" command and its associated commands are used to define the authors and their affiliations.

\author{Fengjie Chang}
\affiliation{%
	\institution{The State Key Laboratory of Networking and Switching Technology}
	\institution{Beijing University of Posts and Telecommunications}
	\city{Beijing}
	\country{China}
}
\email{changfengjie@bupt.edu.cn}

\author{Xinning Zhu}
\affiliation{%
	\institution{Beijing University of Posts and Telecommunications}
	\city{Beijing}
	\country{China}
}
\email{zhuxn@bupt.edu.cn}

\author{Zheng Hu}
\authornote{corresponding Author.}
\affiliation{%
	\institution{Beijing University of Posts and Telecommunications}
	\city{Beijing}
	\country{China}
}
\email{huzheng@bupt.edu.cn}

\author{Yang Qin}
\affiliation{%
	\institution{Beijing University of Posts and Telecommunications}
	\city{Beijing}
	\country{China}
}
\email{qinyoung@bupt.edu.cn}

%%
%% The abstract is a short summary of the work to be presented in the
%% article.
\begin{abstract}
Trajectory User Linking (TUL), which links anonymous trajectories with users who generate them, plays a crucial role in modeling human mobility. Despite significant advancements in this field, existing studies primarily neglect the high-order inter-trajectory relationships, which represent complex associations among multiple trajectories, manifested through multi-location co-occurrence patterns emerging when trajectories intersect at various Points of Interest (POIs). Furthermore, they also overlook the variable influence of POIs on different trajectories, as well as the user class imbalance problem caused by disparities in user activity levels and check-in frequencies.
To address these limitations, we propose a novel HyperGraph-based multi-perspective Trajectory User Linking model (HGTUL). Our model learns trajectory representations from both relational and spatio-temporal perspectives: (1) it captures high-order associations among trajectories by constructing a trajectory hypergraph and leverages a hypergraph attention network to learn the variable impact of POIs on trajectories; (2) it models the spatio-temporal characteristics of trajectories by incorporating their temporal and spatial information into a sequential encoder. Moreover, we design a data balancing method to effectively address the user class imbalance problem and experimentally validate its significance in TUL. Extensive experiments on three real-world datasets demonstrate that HGTUL outperforms state-of-the-art baselines, achieving improvements of 2.57\%\textasciitilde20.09\% and 5.68\%\textasciitilde26.00\% in ACC@1 and Macro-F1 metrics, respectively.
\end{abstract}

\maketitle

%%% do not modify the following VLDB block %%
%%% VLDB block start %%%
\pagestyle{\vldbpagestyle}
\begingroup\small\noindent\raggedright\textbf{PVLDB Reference Format:}\\
\vldbauthors. \vldbtitle. PVLDB, \vldbvolume(\vldbissue): \vldbpages, \vldbyear.\\
\href{https://doi.org/\vldbdoi}{doi:\vldbdoi}
\endgroup
\begingroup
\renewcommand\thefootnote{}\footnote{\noindent
This work is licensed under the Creative Commons BY-NC-ND 4.0 International License. Visit \url{https://creativecommons.org/licenses/by-nc-nd/4.0/} to view a copy of this license. For any use beyond those covered by this license, obtain permission by emailing \href{mailto:info@vldb.org}{info@vldb.org}. Copyright is held by the owner/author(s). Publication rights licensed to the VLDB Endowment. \\
\raggedright Proceedings of the VLDB Endowment, Vol. \vldbvolume, No. \vldbissue\ %
ISSN 2150-8097. \\
\href{https://doi.org/\vldbdoi}{doi:\vldbdoi} \\
}\addtocounter{footnote}{-1}\endgroup
%%% VLDB block end %%%

%%% do not modify the following VLDB block %%
%%% VLDB block start %%%
\ifdefempty{\vldbavailabilityurl}{}{
\vspace{.3cm}
\begingroup\small\noindent\raggedright\textbf{PVLDB Artifact Availability:}\\
The source code, data, and/or other artifacts have been made available at \url{https://github.com/changfengjie3003/HGTUL}.
\endgroup
}
%%% VLDB block end %%%

\section{Introduction}

In recent years, with the rapid development of GPS-based devices and Location-Based Social Networks (LBSN), vast amounts of spatio-temporal trajectory data have been recorded and stored. These data not only contain rich geographical information but also deeply reflect human daily activities and mobility patterns. Through the analysis of trajectory data, researchers can gain insights into complex issues in fields such as urban traffic optimization \cite{tdrive}, population flow dynamics \cite{human}, and personalized recommendation systems \cite{deepmove}. Among these, \textbf{Trajectory User Linking (TUL)} was recently proposed by Gao et al \cite{TULER}. The core objective of TUL is to link the anonymous trajectories to the users who generate them. This task holds significant application value in both public safety and commercial fields. For instance, in public safety, TUL can assist in identifying the movement patterns of potential terrorists or criminal suspects; in the commercial domain, it can support personalized recommendations by analyzing user behavior patterns.

%In the TUL task, the key challenge lies in capturing the movement patterns within trajectory data and linking them to users. association

Over the past few years, substantial efforts have been devoted to solving the TUL problem. Deep learning-based methods for TUL can be primarily categorized into two main approaches. The \textbf{\textit{sequence-based models}} \cite{TULER} \cite{TULVAE}\cite{deeptul} capture user movement patterns by analyzing the time-series features of trajectory points. TULER ~\cite{TULER} first utilizes word embedding \cite{wordemb} to learn location representations, which are then fed into an RNN model to capture sequential transition patterns for TUL. DeepTUL ~\cite{deeptul} employ an attentive RNN to learn from labeled historical trajectory, capturing the multi-periodicity of human mobility while alleviating data sparsity.
The \textbf{\textit{graph-based models}} \cite{gnntul}\cite{s2tul}\cite{attntul} leverage GNNs to model locations or trajectories for TUL. GNNTUL \cite{gnntul} builds a check-in graph that combines geographical and temporal information, utilizing a GNN to model user mobility patterns effectively. S2TUL \cite{s2tul} is the first attempt to incorporate trajectory-level information for modeling the TUL problem. It models the inter-trajectory relationships via multiple homogeneous graphs and a GCN for TUL.

Despite significant prior research, several key challenges still persist in the TUL problem. 

\begin{figure}[t]
	\centering
	% 左侧放置子图a，调整宽度确保居中
	\begin{minipage}[c]{0.65\linewidth} % 调整左侧宽度
		\centering
		\begin{subfigure}[t]{\linewidth}
			\includegraphics[width=\linewidth]{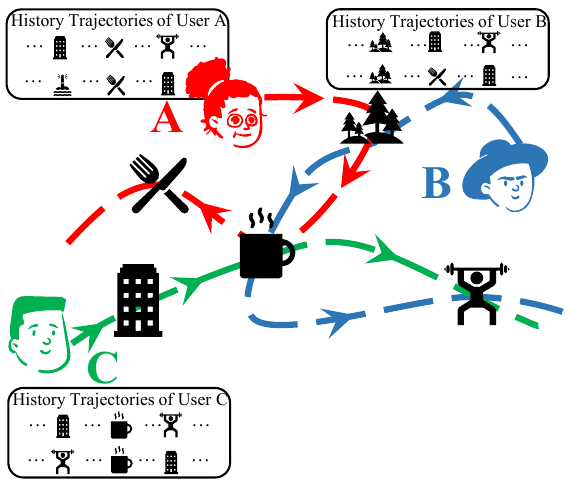}
			\caption{Trajectories Among Users.}
			\label{fig:trajectory}
		\end{subfigure}
	\end{minipage}
	%\hspace{0.05\linewidth} % 增加间距，确保右侧子图不紧挨
	% 右侧放置子图b和c，垂直排列
	\begin{minipage}[c]{0.34\linewidth} % 调整右侧宽度
		\centering
		\begin{subfigure}[t]{\linewidth}
			\includegraphics[width=\linewidth]{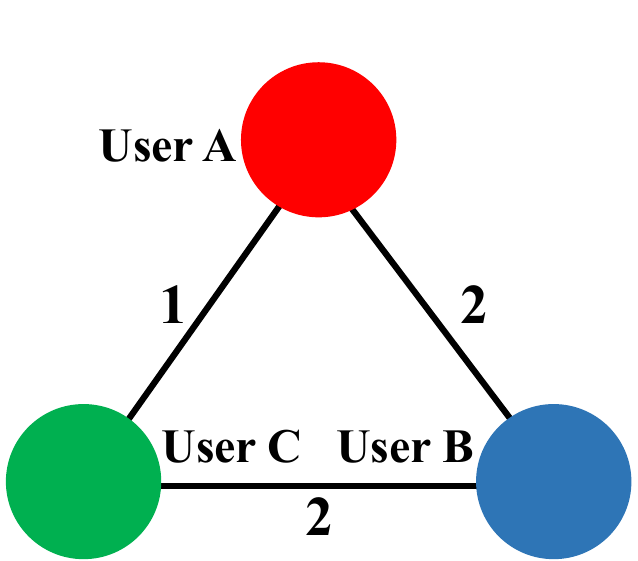}
			\caption{Trajectory Graph.}
			\label{fig:graph}
		\end{subfigure}
		
		%\vspace{0.5em} % 调整b和c之间的垂直间距
		
		\begin{subfigure}[t]{\linewidth}
			\includegraphics[width=\linewidth]{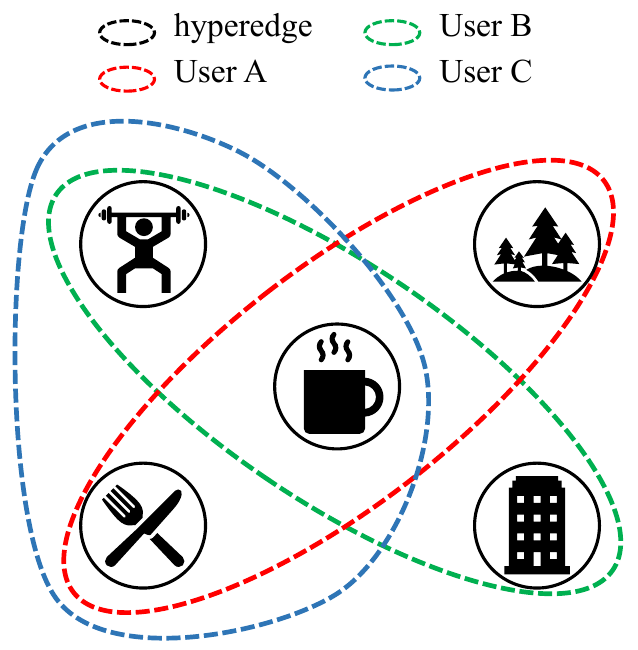}
			\caption{Trajectory Hypergraph.}
			\label{fig:hypergraph}
		\end{subfigure}
	\end{minipage}
	\vspace{-1.5em}
	\caption{Modeling Association                            Relationships Among Trajectories.}
	\label{fig:association_relationships}
	\vspace{-0.85cm}
\end{figure}

The most critical challenge in TUL lies in insufficiently modeling the high-order inter-trajectory relationships (\textbf{\textit{Challenge I}}). High-order inter-trajectory relationships represent complex associations among multiple trajectories, manifested through multi-location co-occurrence patterns emerging when multiple trajectories intersect at various \textbf{Points of Interest (POIs)}. As illustrated in \autoref{fig:trajectory}, Users A, B, and C all visit a café, while Users A and B both visit a park, and Users B and C both visit a gym. Traditional pairwise graph models, which represent connections solely through nodes and edges, fail to accurately capture these multi-location co-occurrence relationships. We refer to these relationships as high-order inter-trajectory associations. Effectively modeling such associations is crucial for TUL task, as it uncovers hidden relationships between users based on shared behaviors, offering a more comprehensive understanding of user mobility patterns.

However, in existing methods for TUL, \textbf{\textit{sequence-based models}} primarily focus on time-series features of trajectories but often neglect the complex associations among them. On th other hand, some \textbf{\textit{graph-based models}} such as S2TUL\cite{s2tul} and AttnTUL\cite{attntul} attempt to address this limitation by integrating trajectory-level information. However, due to the limitations inherent in traditional pairwise graph structures, they are unable to fully capture high-order associations between trajectories.

As shown in \autoref{fig:graph}, traditional graph-based models represent trajectories as nodes, with edges weighted according to the number of overlapping POIs. However, this approach overlooks the POIs themselves and reduce trajectory relationships to simple edge weight. As a result, different trajectory associations may receive the same weight, even if they involve different underlying POIs. For example,  in this graph, the association weight between the trajectories of User B and User A is the same as that between User C and User B (both are 2), despite potential differences in the specific POIs involved. This simplified graph-based representation fails to account for the complex relationships among trajectories that occur at different POIs, leading to the loss of important high-order association information.

In recent years, hypergraphs have gained attention \cite{hypoi}\cite{hguid} for their ability to model high-order relationships. Unlike traditional graphs, which represent connections with pairwise edges, hypergraphs allow a single hyperedge to connect multiple nodes, effectively capturing higher-order relationships. For instance, as shown in \autoref{fig:hypergraph}, a user's trajectory can be represented by a hyperedge, with POIs modeled as nodes within the hypergraph. This approach not only preserves the detailed information of the POIs but also provides a more accurate representation of the higher-order associations between trajectories.

Beyond the core challenge, two additional challenges need to be considered. One of them is the insufficient modeling of the variable impact of POIs on trajectories (\textbf{\textit{Challenge II}}). The influence of a POI on a trajectory is not uniform, as some POIs may have a stronger impact on certain trajectories while being less significant for others. As shown in \autoref{fig:trajectory}, User A's trajectory passes through both a park and a restaurant. However, since the restaurant POI appears more frequently in User A's historical trajectories, it should have a greater impact on this trajectory, as the restaurant POI is more distinctive in identifying A's movement patterns compared to the park POI. Existing methods typically model the relationship between POIs and trajectories using simple, fixed mappings (such as concatenation or summation), without adequately considering users' varying preferences for different POIs. This makes it difficult to accurately capture the variable influence of different POIs on trajectories.

And the other challenge is the neglect of user class imbalance in the data (\textbf{\textit{Challenge III}}). In real-world scenarios, there is significant variation in users' check-in behaviors, with some users being highly active and others check in less frequently. In the case of imbalanced data distribution, models tend to link trajectories to active users, while failing to correctly identify the trajectories of inactive users. However, most existing research has not adequately considered this issue.

To address the aforementioned challenges, we propose a novel HyperGraph-based multi-perspective Trajectory User Linking model (HGTUL), which significantly improves the linking accuracy by introducing a hypergraph structure. First, to tackle the critical issue of insufficiently modeling high-order inter-trajectory relationships (\textbf{\textit{Challenge I}}), we represent trajectories as hyperedges and POIs as vertices, constructing a trajectory hypergraph where multiple trajectories can share common POIs. This structure effectively models the complex high-order associations among trajectories. To mitigate the impact of user class imbalance (\textbf{\textit{Challenge III}}), we employ a simple data balancing method during preprocessing to ensure fairer model leaning across active and inactive users.
%ensuring that the model performs well across different user groups.
Next, to capture the variable impact of POIs on trajectories (\textbf{\textit{Challenge II}}), we utilize a hypergraph attention network to learn POIs' varying impacts, generating attentive trajectory representations. We fuse these attentive representations with structural trajectory representations that capture the inherent structure of trajectories, obtaining trajectory representations from a \textit{\textbf{relational perspective}}. Additionally, to fully leverage the spatio-temporal characteristics of trajectories, we encode temporal and spatial information as sequential features and inputting them into an Long Short-Term Memory (LSTM)\cite{lstm} network to learn trajectory representations from a \textit{\textbf{spatio-temporal perspective}}. Finally, the multi-perspective trajectory representations are fused and input into a classification layer to predict the user who generated the trajectory. 

%Specifically, our approach consists of the following key steps: First, we employ a simple data balancing method to preprocess the input data, thereby mitigating the impact of user class imbalance (addressing \textbf{\textit{Challenge III}}). Second, we regard trajectories and POIs as hyperedges and vertices, constructing a trajectory hypergraph to capture high-order associations among trajectories (addressing \textbf{\textit{Challenge I}}). Building on this, we utilize a hypergraph attention network to learn the variable influence of POIs on trajectories, generating attentive trajectory representations that reflect the varying impacts of POIs (addressing \textbf{\textit{Challenge II}}). Furthermore, we fuse these attentive representations with structured trajectory representations that capture the inherent trajectory structures, obtaining trajectory representations from a relational perspective. Additionally, to fully leverage the spatio-temporal characteristics of trajectories, we encode temporal and spatial information as sequential features, inputting them into an Long Short-Term Memory (LSTM)\cite{lstm} network to learn trajectory representations from a spatio-temporal perspective. Finally, the multi-perspective trajectory representations are fused and input into a classification layer to predict the user who generated the trajectory. 

The contributions of our work are summarized as follows:
%\vspace{-0.5cm}
\begin{itemize}[leftmargin=2em]
	\item We propose a hypergraph-based multi-perspective trajectory user linking model, which integrates high-order inter-trajectory relationships and spatio-temporal characteristics of trajectories. 
	\item We introduce a hypergraph attention network to learn the variable influence of POIs on trajectories, enhancing the accuracy of user linking by learning the key POIs within user trajectories.
	\item We design a data balancing method to effectively enhance the robustness of model and accuracy for inactive users. Furthermore, by applying data balancing to other baseline models, we validate its necessity in TUL.
	\item We conduct extensive experiments on three real-world datasets, demonstrating that our model significantly outperforms state-of-the-art baselines in terms of ACC@1 and Macro-F1, with improvements ranging from 2.57\% to 20.09\% and 5.68\% to 26.00\%, respectively.
\end{itemize}

\section{Related Work}

\subsection{Trajectory-User Linking}

Trajectory data provides unprecedented insights into human mobility patterns. Recently, TUL problem was introduced in ~\cite{TULER},  which links trajectories to their generating-users, and gradually becomes a hot topic in spatio-temporal data mining. Deep learning-based methods for TUL can be broadly categorized into two distinct modeling approaches:

(1) \textbf{\textit{Sequence-based modeling methods}} represent trajectories as time-series to address the TUL task. TULER ~\cite{TULER} utilizes RNNs to learn sequential transition patterns from trajectory data and link them to users. It first employs word embedding to learn location representations and then fed them into RNN model to capture user mobility patterns for TUL. However, standard RNN-based models face data sparsity issues due to their limitation in leveraging the unlabeled data that inherently contains rich information about user mobility patterns. In their subsequent work ~\cite{TULVAE}, TULVAE alleviates the data sparsity problem by leveraging large-scale unlabeled data and captures the hierarchical and structural semantics of trajectories through \textbf{Variational Autoencoder (VAE)}. However, it does not exploit the rich features within trajectories or consider the multi-periodicity of human mobility. DeepTUL ~\cite{deeptul} addresses this limitation by employing an attentive recurrent network to learn from historical trajectory, capturing the multi-periodicity of human mobility while mitigating data sparsity. However, these methods primarily rely on existing sequence models such as LSTM ~\cite{TULER}~\cite{TULVAE} or attention mechanism \cite{deeptul} to capture intra-trajectory information and generate trajectory representations but fail to capture the global association relationships between trajectories.

(2) \textbf{\textit{Graph-based methods}} leverage GNNs to model locations or trajectories, capturing more complex and diverse relationships. Instead of only relying on visited sequences as previous methods did, GNNTUL \cite{gnntul} constructs a check-in graph that integrates geographical and temporal information and leverages GNN to effectively model user mobility patterns. However, despite modeling transition patterns using GNN, it remains limited in capturing inter-trajectory relationships.
S2TUL \cite{s2tul} is the first attempt to incorporate trajectory-level information for modeling the TUL problem. It models the complex relationships between trajectories by constructing multiple homogeneous and heterogeneous graphs, and then passes this information through a GCN to learn trajectory representations. And then, it combines intra-trajectory and inter-trajectory information to predict the generating users of trajectories.
%Contemporaneously, ANES \cite{aspect} focuses on representation learning for social link inference based on user trajectory data. It leverages user trajectory data and bipartite graphs to learn aspect-oriented relationships between users and POIs, thereby better capturing user behavioral patterns.
Nevertheless, the aforementioned methods overlook the integration and synergy of local and global information. AttnTUL \cite{attntul} proposes a hierarchical spatio-temporal attention neural network, which simultaneously models the local and global spatio-temporal characteristics of user mobility trajectories through a GNN architecture. And it designs a hierarchical attention network to jointly encode local transition patterns and global spatial dependencies for TUL. However, due to the limitations of traditional graph structures, they can only model pairwise relationships and fail to effectively capture high-order inter-trajectory association relationships, thus limiting the representation of mobility patterns in trajectories.

\subsection{Hypergraph Learning}

Hypergraph \cite{hycn}\cite{hygnn}\cite{hgnn+} is a generalization of graph, which can naturally model complex higher-order relationships among vertices by connecting multiple vertices simultaneously through hyperedges. Due to its remarkable ability to capture higher-order relations, hypergraphs have increasingly drawn significant attention from researchers. To efficiently learn deep embeddings on higher-order graph-structured data, \cite{hycn} proposed to extend the graph neural network architecture to hypergraphs by introducing two end-to-end trainable operators, namely hypergraph convolution and hypergraph attention.

In recent years, hypergraph neural networks have been widely used in spatio-temporal data modeling and user mapping tasks. In user mapping, UMAH \cite{hguid} models social structure and user profile relationships in a unified hypergraph, learns a common subspace by preserving the hypergraph structure as well as the correspondence relations of labeled users, and facilitates user mapping across social networks based on similarities in the subspace. For spatio-temporal data modeling, STHGCN \cite{hypoi} constructs a hypergraph to model trajectory granularity information, and captures higher-order information including collaborative relationships between trajectories through hypergraph learning. These studies highlight the significant potential of hypergraph-based learning methods and offer a novel solution approach to the trajectory user linking problem.

\section{METHODOLOGY}

\begin{figure*}
		\centering
		\includegraphics[width=\textwidth]{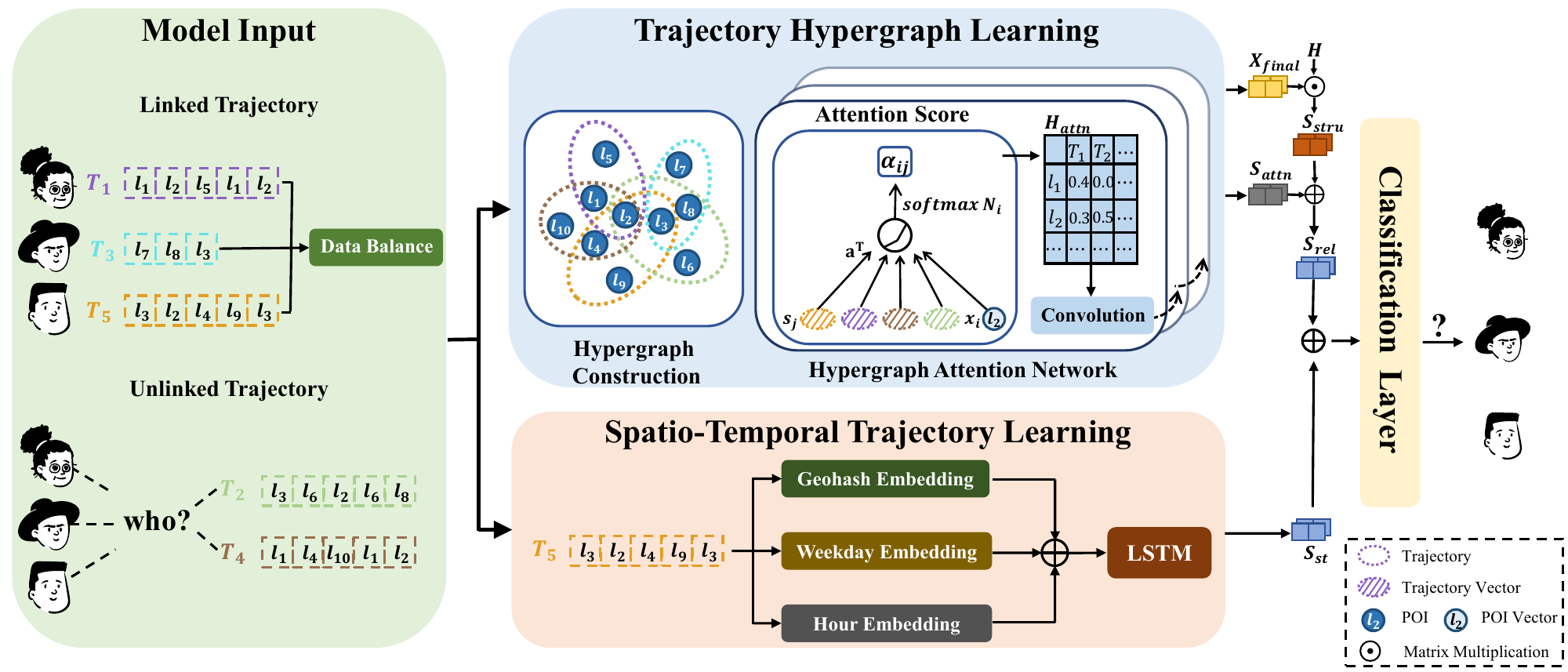} % 使用 \textwidth 占据双栏宽度
		\caption{The proposed framework.}
		\label{fig:framework}
	\end{figure*}

To address the limitations of existing research, we propose \textbf{HGTUL}, a HyperGraph-based multi-perspective Trajectory-User Linking model, as shown in \autoref{fig:framework}. The framework consists of three main components: (1) trajectory hypergraph learning, (2) spatio-temporal trajectory learning, and (3) a classification layer.
\textit{First}, we construct a trajectory hypergraph to model the higher-order inter-trajectory relationships. Using a hypergraph attention network, we learn trajectory representations from the relational perspective. \textit{Second}, we utilize sequence modeling to learn trajectory representations from a spatio-temporal perspective. \textit{Finally}, we fuse the relational and spatio-temporal trajectory representations to classify trajectories by their users.
	
\subsection{Preliminary}
We first introduce some preliminary concepts and define the TUL problem.

\textit{Definition1 Spatio-Temporal Point.} A spatio-temporal point \( p = (lon, lat, t) \) represents a visit to a POI at coordinates \( (lon, lat) \) at time \( t \). The set of all POIs is denoted as \( \mathcal{P} = \{(lon_1, lat_1), (lon_2, lat_2), \allowbreak..., (lon_L, lat_L)\} \), where $L$ is the number of POIs.

\textit{Definition2 Linked Trajectory}. A linked trajectory is a sequence of spatio-temporal points generated by a user $u_{i}$ during a certain time interval, denoted as $T_{u_{i}}=\{p_{i_1},p_{i_2},...,p_{i_n}\}$. The set of all linked trajectories is represented as $\mathcal{T}_{u} = \{T_{u_1}, T_{u_2}, ..., T_{u_Q}\}$, where $Q$ is the total number of users.

\textit{Definition3 Unlinked Trajectory}. Let $T_{k}=\{p_{k_1},p_{k_2},...,p_{k_n}\}$ denote a trajectory with an unknown generating user, called an \textit{unlinked trajectory}. The set of all unlinked trajectories is represented as $\overline{\mathcal{T}} = \{T_{k_1}, T_{k_2}, ..., T_{k_w}\}$, where $w$ is the number of unlinked trajectories.

\textit{Definition4 Hypergraph}. A hypergraph can be represented by $\mathcal{G} = (\mathcal{V}, \mathcal{E})$, where $\mathcal{V}$ is the set of vertices with size $|\mathcal{V}|$ and $\mathcal{E}$ is the set of hyperedges with size $|\mathcal{E}|$. A hyperedge $e \in \mathcal{E}$ can connect any number of vertices $v \in \mathcal{V}$. The incidence matrix $\mathbf{H} \in \mathbb{R}^{|\mathcal{V}| \times |\mathcal{E}|}$ is used to represent the topological structure of the hypergraph. It is defined such that ${H}_{i,j} = 1$ if $v_i \in e_j $ , ${H}_{i,j} = 0$ otherwise.  

\textit{TUL Problem Formulation}. Given the set of unlinked trajectories $\overline{\mathcal{T}}$ generated by users $\mathcal{U} = \{u_1, u_2, \dots, u_Q\}$ and a corresponding set of linked trajectories $\mathcal{T}_{u}$. The goal of \textit{TUL} is to learn a mapping function $f:\overline{\mathcal{T}} \to \mathcal{U}$ that links unlinked trajectories with their respective users.
%\vspace{-1em}
\subsection{Trajectory Hypergraph Learning}
\textit{3.2.1 Trajectory Hypergraph Construction.} 
%To capture the high-order relationships between POIs and the global associations among trajectories, we construct a trajectory hypergraph $\mathcal{G} = (\mathcal{V}, \mathcal{E})$. As shown in Figure 1, the vertices of trajectory hypergraph represent POIs, with each trajectory modeled as a hyperedge, i.e., $\mathcal{V}\equiv\mathcal{P}$ and $\mathcal{E}\equiv\mathcal{T}\cup\overline{\mathcal{T}}$. The hyperedge property intuitively characterizes the high-order relationships between POIs. Additionally, to describe the interactions between trajectories and POIs, we construct an incidence matrix $\mathbf{H} \in \mathbb{R}^{L \times N}$, where $L$ is the number of POIs and  $N$ is the number of all trajectories(including both linked and unlinked trajectories, with $N > w$). 
To capture the high-order association relationships among trajectories, we construct a trajectory hypergraph $\mathcal{G} = (\mathcal{V}, \mathcal{E})$. As shown in \autoref{fig:framework}, the vertices of trajectory hypergraph represent POIs, with each trajectory modeled as a hyperedge, i.e., $\mathcal{V}=\mathcal{P}$ and $\mathcal{E}=\mathcal{T}\cup\overline{\mathcal{T}}$. Additionally, to describe the interactions between trajectories and POIs, we construct an incidence matrix $\mathbf{H} \in \mathbb{R}^{L \times N}$, where $L$ is the number of POIs and  $N$ is the number of all trajectories (including both linked and unlinked trajectories). The hyperedge property intuitively characterizes the complex relationships between POIs. By representing multiple trajectories as hyperedges that share vertices, the hypergraph can effectively model the complex high-order associations among these trajectories.

Leveraging the trajectory hypergraph, the co-occurrence relationships among POIs facilitate the propagation of information across trajectories. For instance, POIs overlapping with multiple trajectories can reveal latent patterns in visitation behaviors. The trajectory hypergraph not only captures the POI information but also reveals individual behavior patterns through high-order trajectory associations, thus offering a relational perspective for TUL.
\vspace{0.2cm} %

\noindent\textit{3.2.2 Trajectory Hypergraph Attention Network.} After constructing the trajectory hypergraph, we introduce the trajectory hypergraph attention network to learn POI representations and the relational trajectory representation based on them. Before encoding, the POI embeddings $\mathbf{P}\in\mathbb{R}^{L\times d}$ are initialized via a look-up table, where $d$ denotes the embedding dimension.
%Before encoding, the trajectory embeddings $\mathbf{S}\in\mathbb{R}^{N\times d}$ and POI embeddings $\mathbf{P}\in\mathbb{R}^{L\times d}$ are initialized via a look-up table, where $d$ denotes the embedding dimension.

POIs visited within the same trajectory exhibit stronger transitional dependencies compared to randomly paired POIs. We employ hypergraph convolution to model these dependencies. Based on the hypergraph convolution operator in \cite{hycn}, the layer-wise propagation process in the trajectory hypergraph convolution network can be expressed as: 
\begin{equation}
	\mathbf{X}^{(m+1)}=\sigma(\mathbf{D}^{-1/2}\mathbf{H}\mathbf{B}^{-1}\mathbf{H}^{\mathrm{T}}\mathbf{D}^{-1/2}\mathbf{X}^{(m)}\mathbf{W}^{(m)})
	\label{eq:hypconv}
\end{equation}
where $\mathbf{W}^{(m)}$ is the layer-specific learnable weight matrix, with $m$ denoting the layer index in the trajectory hypergraph convolution network. $\mathbf{H}$ is the hypergraph incidence matrix. $\mathbf{D}$ is the vertices degree matrix that be calculated as $D_{ii} = \sum_{j=1}^{|\mathcal{E}|} H_{ij}$. $\mathbf{B}$ is the hyperedge degree matrix that be calculated as $B_{jj} = \sum_{i=1}^{|\mathcal{V}|} H_{ij}$. The term $\mathbf{D}^{-1/2}\mathbf{H}\mathbf{B}^{-1}\mathbf{H}^{\mathrm{T}}\mathbf{D}^{-1/2}$ performs symmetric normalization to balance information aggregation and stabilize computation. $\mathbf{X}^{(m)}\in\mathbb{R}^{L\times d}$ is the output of $m$-th layer, where $d$ is the embedding dimension, and $\mathbf{X}^{(0)}=\mathbf{P}$. $\sigma$ is a non-linear activation function. 

However, the relationship between trajectories and POIs is variable rather than fixed. From the perspective of trajectories contributing to POI representation, not all trajectories contribute equally to a POI representation. For example, if a user visits a POI only occasionally (e.g., User A visiting the park POI in \autoref{fig:trajectory}), her trajectories have a weaker influence on the POI's representation compared to trajectories of users who visit the POI frequently (e.g., User B visiting the park POI in \autoref{fig:trajectory}). Conversely, from the perspective of POI contributing to trajectories representation, POIs that a user visits frequently (e.g., the restaurant POI of User A in \autoref{fig:trajectory}) play a more significant role in distinguishing the user and characterizing their trajectories. Thus, it is crucial to capture this variable relationship in TUL.

To effectively model the variable relationship, we introduce an attention learning module into the trajectory-POI incidence matrix $\mathbf{H}$. The module dynamically adjusts the incidence matrix by computing attention weights between trajectories and POIs. The trajectory embedding $\mathbf{s}_{j} \in \mathbb{R}^{d}$ is initialized as a learnable vector and updated during training to capture the complex relationships between POIs and trajectories. 
For a given POI $p_{i}$ with its embedding $\mathbf{x}_{i}$, and its associated trajectory hyperedge $T_{j}$ with its embedding $\mathbf{s}_{j}$, the similarity\cite{hycn} between POI $p_{i}$ and trajectory $T_{j}$ can be measured as
\begin{equation}
	\mathrm{sim}(\mathbf{x}_{i},\mathbf{s}_{j})=\mathbf{a}^\mathrm{T}[\mathbf{x}_{i}\|\mathbf{s}_{j}]
\end{equation}
Here $\|$ denotes concatenation and $\mathbf{a}$ is a learnable weight vector used to output a scalar similarity value. 

And the attention score between them can be calculated as 
\begin{equation}
	H_{ij}^{attn}=\frac{\exp\left(\sigma\left(\mathrm{sim}(\mathbf{x}_{i},\mathbf{s}_{j})\right)\right)}{\sum_{k\in\mathcal{N}_i}\exp\left(\sigma\left(\mathrm{sim}(\mathbf{x}_{i},\mathbf{s}_{k})\right)\right)}
	\label{eq:attn}
\end{equation}
A non-linear activation function $\sigma$ is applied to enhance the expressiveness of the similarity.  $\mathcal{N}_i$ is the neighborhood set of POI $p_{i}$, which consists of the trajectory hyperedges containing $p_{i}$. 
These scores populate the attentive incidence matrix $\mathbf{H}_{attn}$, which has the same size $L \times N$ as the incidence matrix $\mathbf{H}$. Each element in $\mathbf{H}_{attn}$ represents the attention score between $p_{i}$ and $T_{j}$, capturing their variable relationships.

The trajectory embeddings $\{\mathbf{s}_{j}\}_{j=1}^{N}$ are initialized as learnable vectors and organized into a matrix $\mathbf{S}_{attn} \in \mathbb{R}^{N \times d}$, where each row represents the embedding of a specific trajectory $T_{j}$. During training, $\mathbf{s}_{j}$ is dynamically updated based on the attention score $H_{ij}^{attn}$, which quantify the importance of each POI $p_{i}$ to trajectory $T_{j}$. This allows $\mathbf{S}_{attn}$ to capture the variable influence of POIs on the trajectories, making it an effective attentive representation of all trajectories.
%And $\mathbf{x}_{i}$ is the learnable embedding vector of $p_{i}$, $\mathbf{s}_{j},\mathbf{s}_{k}$ are the learnable embedding vectors of trajectory $T_{j},T_{k}$. And the attentive trajectory representation can also be updated as $\mathbf{S}_{attn}\in\mathbb{R}^{N\times d}$,where $d$ is the embedding dimension.

With the attentive incidence matrix $\mathbf{H}_{attn}$ enriched by the attention module, one can also follow \autoref{eq:hypconv} to learn the layer-wise attentive embeddings of POIs.
\begin{equation}
\begin{aligned}
	\mathbf{X}_{attn}^{(m+1)}&=\mathbf{AT}\left(\mathbf{X}_{attn}^{(m)}\right) \\ &=\sigma(\mathbf{D}^{-1/2}\mathbf{H}_{attn}^{(m)}\mathbf{B}^{-1}\mathbf{H}_{attn}^{(m)\mathrm{T}}\mathbf{D}^{-1/2}\mathbf{X}_{attn}^{(m)}\mathbf{W}_{attn}^{(m)})
\end{aligned}
	\label{eq:attpoi}
\end{equation}
where $\mathbf{H}_{attn}^{(m)}$ is the $m$-th attentive incidence matrix.

We employ dropout and residual connections to address overfitting and over-smoothing. The final embeddings of POIs are obtained by averaging the embeddings from all layers.
\begin{equation}
	\mathbf{X}_{attn}^{(m+1)}  =\mathrm{Dropout}\left(\mathbf{AT}\left(\mathbf{X}_{attn}^{(m)}\right)\right)+\mathbf{X}_{attn}^{(m)}
\end{equation}
\begin{equation}
	\mathbf{X}_{\mathrm{final}}=\frac{1}{M+1}\sum_{l=0}^M\mathbf{X}_{attn}^{(m)}
\end{equation}
where $M$ denotes the total number of trajectory hypergraph attention network layers. And $\mathbf{AT( )}$ propagates POI features through hypergraph attention, as defined in \autoref{eq:attpoi}. $\mathbf{X}_{\mathrm{final}}\in\mathbb{R}^{L\times d}$ is the final embeddings of POIs.
%The trajectory is a sequence of POIs, and this structure naturally captures the essential information of the trajectory. Thus, we compute its structural representation as:

A trajectory is naturally represented as a collection of visited POIs, reflecting its compositional structure. To capture this property, we compute its structural representation by summing the embeddings of all POIs in the trajectory:
\begin{equation}
	\mathbf{S}_{stru}=\mathbf{X}_{\mathrm{final}}\cdot\mathbf{H}
\end{equation}
\begin{equation}
	\mathbf{S}_{rel} =\mathbf{S}_{attn}+\mathbf{S}_{stru}
	\label{eq:globaltra}
\end{equation}
where the $(\cdot)$ represents matrix multiplication, and $\mathbf{H}$ denotes the original incidence matrix without attention calculations. $\mathbf{S}_{stru}\in\mathbb{R}^{N\times d}$ represents the structural trajectory representation, where $N$ is the number of all trajectories. Finally, the relational representation of the trajectory is obtained by combining the attentive trajectory representation and the structural representation, as shown in \autoref{eq:globaltra}. The proposed trajectory hypergraph attention network learns relational trajectory representations while capturing both POI information and high-order inter-trajectory relationships.

\subsection{Spatio-temporal Trajectory Learning}
In the aforementioned trajectory hypergraph learning module, we capture the relational trajectory representations based on hypergraph. However, the temporal and spatial information of visited POIs has not been fully leveraged. So we introduce a sequence modeling module to enhance trajectory representation from a spatio-temporal perspective.

\noindent\textit{3.3.1 Spatio-temporal Encoder.} To capture the geographical information of POIs, we encode the raw POI coordinates by GeoHash \cite{geohash}. GeoHash is a grid-based geographical encoding method that maps continuous latitude and longitude coordinates into a discrete string grid, effectively representing the geographic location of a POI. For a given POI with original coordinates $(lon_k, lat_k)$, we encode it as  $g_k = GeoHash(lon_k, lat_k)$, and the corresponding spatial embedding $\mathbf{g}_{ek} = Embedding(g_k)\in\mathbb{R}^{d} $.

To model the temporal information of POI visits, we divide time into two granularities: first, we partition the day into 48 time slots with a 0.5-hour interval. Second, we divide the time into 2 categories  based on whether it is a weekday or not. For a given POI visit time $t_k$, we generate the corresponding hour-level embedding $\mathbf{t}_{hk}\in\mathbb{R}^{d} $ and week-level embedding $\mathbf{t}_{wk}\in\mathbb{R}^{d} $ based on this time partitioning.

Then we obtain the embedding $\mathbf{x}_{st\textunderscore k}\in\mathbb{R}^{d}$ for spatio-temporal point $p_k$ by summing its spatial and temporal embeddings as follows:
\begin{equation}
	\mathbf{x}_{st\textunderscore k}=\mathbf{g}_{ek}+\mathbf{t}_{hk}+\mathbf{t}_{wk}
\end{equation}
where $d$ is the embedding dimension. Finally, the initial spatio-temporal trajectory representation $\mathbf{S}_{ist\textunderscore i}$ is constructed by concatenating the embeddings of spatio-temporal points as follows:
\begin{equation}
	\mathbf{S}_{ist\textunderscore i} = \mathbf{x}_{st\textunderscore 1} \| \ldots \| \mathbf{x}_{st\textunderscore |T|}
\end{equation}
where $	\mathbf{S}_{ist\textunderscore i}\in \mathbb{R}^{|T| \times d}$, and $|T|$ is the length of the trajectory.
\vspace{0.2cm}

\noindent\textit{3.3.2 Sequence Modeling with LSTM.}
Since trajectory hypergraph learning has captured the relational information, we focus on the sequential features in this perspective. To ensure model efficiency, we choose LSTM \cite{lstm} for modeling the spatio-temporal characteristics of the trajectories, rather than more complex models like Transformer \cite{transformer} or other recent time series modeling architectures \cite{mamba}. Specifically, we use the hidden state at the final timestep of the LSTM to represent the trajectory:
\begin{equation}
	\mathbf{S}_{st\textunderscore i} = \mathbf{LSTM}(\mathbf{S}_{ist\textunderscore i})
\end{equation}
where $\mathbf{S}_{st\textunderscore i}\in \mathbb{R}^{d}$ is the spatio-temporal trajectory representation of trajectory $T_i$.

\subsection{Classification Layer}

To link trajectories to their respective users, we obtain the final trajectory representation by summing the relational representation learned from the trajectory hypergraph and the spatio-temporal representation extracted by the LSTM. Both representations are normalized using L2 normalization to ensure consistency in their scale.  The final representation is then fed into a fully-connected layer for user linking. The process is summarized as follows:
\begin{equation}
	\mathbf{S}_{final\textunderscore i} = \mathbf{NORM}(\mathbf{S}_{st\textunderscore i}) + \mathbf{NORM}(\mathbf{S}_{rel\textunderscore i})
\end{equation}
\begin{equation}
\mathit{y_i}=\mathbf{W_c}\mathbf{S}_{final\textunderscore i}+\mathbf{b_c}
\end{equation}
where $\mathbf{NORM}$ denotes L2 normalization, $\mathbf{W_c}\in \mathbb{R}^{Q \times d}$ and $\mathbf{b_c}\in \mathbb{R}^{Q}$ are the learnable parameters of classification layer. $Q$ is the number of users, and $\mathit{y_i}\in \mathbb{R}^{Q}$ is the estimated probability of users for trajectory $T_i$.

We train the parameters of model by minimizing the cross-entropy loss function as follows:
\begin{equation}
\mathcal{L}=-\frac{1}{N_{t}}\sum_{i=1}^{N_{t}}{c_{i}}\log\left(Softmax\left(\mathit{y}_{i}\right)\right)
\end{equation}
where $c_{i}$ is the ground truth label of trajectory $T_i$,  $N_{t}$ is the number of training trajectories.

\subsection{Data Balancing}
TUL is inherently a multi-class classification problem. However, in real-world check-in data, the diversity of user behaviors leads to an imbalanced distribution of trajectory counts across users. To mitigate the impact of data imbalance on model performance, we propose a data balancing method applied to the training set before model training. First, we calculate the average number of trajectories per user in the training set as \( N_{ave} = \frac{N_{t}}{Q} \), where \( N_{t} \) is the number of training trajectories and \(Q\) is the number of users. Next, for users with $N_u < N_{ave}$, $N_u$ is the trajectories number of user $u$ in training set, we randomly replicate their trajectories until the count reaches $N_{ave}$. Additionally, we set a threshold \( \theta _t\). For users with \( N_u > (1 + \theta_t) \times N_{ave} \), we randomly remove trajectories to ensure \( N_u \leq (1 + \theta_t) \times N_{ave} \). By effectively balancing the trajectory distribution across users, this approach improves model performance, particularly for users with fewer trajectories.

\begin{table}[t]
	\centering
	\caption{Statistics of datasets.}
	\label{tab:dataset_statistics}
	\begin{tabular}{lccccc}
		\toprule
		Dataset & Users & Train/Valid/Test & POIs & Length \\ 
		\midrule
		NYC & 500 & 3807/1331/1540 & 4066 & [1, 107] \\ 
		& 1000 & 7563/2650/3034 & 5015 & [1, 107] \\
		JKT & 500 & 4264/1499/1682 & 4451 & [1, 90] \\ 
		& 1000 & 8540/2982/3363 & 5879 & [1, 63] \\
		Gowalla & 500 & 2846/1027/1210 & 3172 & [1, 124] \\ 
		& 1000 & 6045/2166/2507 & 3822 & [1, 128] \\
		\bottomrule
	\end{tabular}
\end{table}
\section{EXPERIMENTS}
In this section, we conduct experiments on three real-world datasets and report the results. The experiments aims to answer the following research questions: \textbf{Q1} How does our proposed model perform compared to existing representative models? \textbf{Q2} How does each module affect the overall performance of the model? \textbf{Q3} How does our model perform in addressing the user cold-start problem? \textbf{Q4} How does data balancing affect the TUL task?
\subsection{Experiment Settings}

\textit{4.1.1 Datasets.} To evaluate the performance of our proposed model, we conduct experiments using three public real-world datasets derived from the raw check-in data of Gowalla\cite{gowalla}, Foursquare\cite{forsqu} in New York City (NYC), and Jakarta (JKT). To ensure the quality of the trajectory sequences, we apply the same filtering rules as \cite{cacsr} to clean the raw data. We filter out users with fewer than 10 check-in records as well as locations visited less than 10 times. Similar to previous studies \cite{TULER}\cite{deeptul}, to simulate the TUL, we sort the cleaned check-in data by time for each user and then segment the trajectories with a one-week time interval. To evaluate the robustness of our model, we randomly select 500 and 1000 users from each dataset for experimentation. The trajectory data for each user is divided into training, validation, and test sets in a 6:2:2 ratio. The statistical information of the preprocessed trajectory data is summarized in \autoref{tab:dataset_statistics}, where Length represents the range of the number of spatio-temporal points in the trajectories of the dataset.
\vspace{0.2cm}

\noindent\textit{4.1.2 Evaluation Metrics.} We evaluate the performance of the proposed model using four key metrics: \textbf{ACC@k}, \textbf{Macro-P}, \textbf{Macro-R}, and \textbf{Macro-F1}. These metrics are commonly used in multi-class classification problems, and TUL can be regarded as such a problem. 
\textbf{ACC@k} represents the rate of whether the true user is included in the Top-k predicted users. It can be calculated as:
\begin{equation}
	\mathbf{ACC@k} = \frac{|\{T_{i} \in \overline{T} : u_{i} \in Top_k(\mathbf{y}_{i})\}|}{|\overline{T}|}
\end{equation}
where $u_{i}$ is the ground truth user, and $Top_k(\mathbf{y}_{i})$ is the predicted Top-k users set.

\textbf{Macro-F1} comprehensively reflects the model's performance across all classes by integrating both precision and recall, and it is widely used in multi-class classification problems.It is can be defined as:
\begin{equation}
	\mathbf{Macro\text{-}R} = \frac{1}{Q} \sum_{i=1}^{Q} R_i\\
\end{equation}
\begin{equation}
	\mathbf{Macro\text{-}P} = \frac{1}{Q} \sum_{i=1}^{Q} P_i\\
\end{equation}
\begin{equation}
	\mathbf{Macro\text{-}F1} = \frac{1}{Q} \sum_{i=1}^{Q} \frac{2 \times P_i \times R_i}{P_i + R_i}	
\end{equation}
where $Q$ is the number of users of unlinked trajectories.
\vspace{0.2cm}

\noindent\textit{4.1.3 Baselines.}
We compare \textbf{HGTUL} with following representative methods for TUL, including (1) Sequence-based methods \textbf{TULER}, \textbf{DeepTUL} and \textbf{CACSR}; (2) Graph-based methods \textbf{S2TUL} and \textbf{AttnTUL}:
\begin{itemize}[leftmargin=2em]
\item \textbf{TULER}~\cite{TULER} It is the original RNN-based approach to learn sequential transition patterns from trajectory data for TUL. There are three variants: RNN with Gated Recurrent Unit \cite{gru} (\textbf{TULER-G}), Long Short-Term Memory \cite{lstm} (\textbf{TULER-L}), and bidirectional LSTM \cite{bilstm} (\textbf{Bi-TULER}).
\item \textbf{DeepTUL}~\cite{deeptul} It is a recurrent network with attention mechanism. It learns from historical trajectory to capture the multi-periodicity of user mobility and mitigate data sparsity for TUL.
\item \textbf{CACSR} \cite{cacsr} It is a representation model for user check-in sequences, built upon bidirectional LSTM, which captures the spatio-temporal features and high-level semantics of check-in sequences through contrastive learning and adversarial perturbations.
\item \textbf{S2TUL} \cite{s2tul} It is the first attempt to incorporate trajectory-level information for TUL. It models the relationships between trajectories by multiple graphs with GCN and combines intra-trajectory and inter-trajectory information for TUL.There are two variants: it instantiates with repeatability graphs(\textbf{S2TUL-R}), heterogeneous graphs formed by merging the repeatability and spatial graphs (\textbf{S2TUL-HRS}).
\item \textbf{AttnTUL} \cite{attntul} It proposes a hierarchical spatio-temporal attention neural network, which models the local and global spatio-temporal characteristics through a GNN architecture and simultaneously encode the intra-trajectory and inter-trajectory dependencies for TUL.
\end{itemize}
\vspace{0.2cm}

\noindent\textit{4.1.4 Parameter Settings.} Our experiments are conducted with PyTorch 1.7.11 on 24GB Nvidia 3090 GPU. For the baselines, we implement them on three real-world datasets using the source code released by the authors and the parameter settings recommended in the original papers. For our \textbf{HGTUL}, we adopt Adam \cite{adam} as the optimizer with a weight decay of 5e-4. The learning rate is dynamically adjusted based on validation performance, with an initial learning rate of 1e-3 and a minimum learning rate of 1e-6. We encode the coordinates of POIs into Geohash codes with 7 characters, where each code represents a grid size of approximately $150m\times150m$. The embedding dimension size is set to $d$=128, the number of hypergraph layers $M$=2, the training epochs is set to 50, the batch size is set to 64, and the dropout rate is set to 0.3. The activation function $\sigma$ is a LeakyReLU with the negative slope of 0.2. The data balancing threshold \( \theta _t\) is set to 0.5. We employ an early stopping mechanism with a patience of 5 to avoid overfitting. We test based on the model parameters that maximize \textbf{ACC@1} on the validation set. For each experiment, we perform 10 repetitions and report the average values along with the standard deviations. All personal information of users in our dataset is anonymized to protect user privacy. Since the study aims to verify the model's performance and robustness, our model is implemented without any parameter tuning.

\subsection{Overall Performance}
\begin{table*}[t]
	\centering
	\caption{Overall Performance of Baselines and Proposed Method In Three Real-World Datasets}
	\label{tab:performance}
	\begin{adjustbox}{width=\textwidth}
	\begin{tabular}{lccccccccccc}
		\toprule
		\multirow{2}{*}{\textbf{Data}} & \multirow{2}{*}{\textbf{Methods}} & \multicolumn{5}{c}{\textbf{Users}=500} & \multicolumn{5}{c}{\textbf{Users}=1000} \\
		\cmidrule(lr){3-7} \cmidrule(lr){8-12}
		& & \textbf{ACC@1(\%)} & \textbf{ACC@5(\%)} & \textbf{Macro-F1(\%)}&\textbf{Macro-P(\%)} & \textbf{Macro-R(\%)} & \textbf{ACC@1(\%)} & \textbf{ACC@5(\%)} & \textbf{Macro-F1(\%)} & \textbf{Macro-P(\%)} & \textbf{Macro-R(\%)} \\
		\midrule
		\multirow{9}{*}{\textbf{GOWALLA}} & \textbf{TULER-L} & $46.55 \pm 0.82$ & $63.10 \pm 0.53$ & $31.96 \pm 0.69$ & $32.27 \pm 0.74$ & $35.22 \pm 0.63$ &38.82 $\pm$ 0.54 & 54.50 $\pm$ 0.68 & 24.97 $\pm$ 0.57 & 25.32 $\pm$ 0.72 & 28.05 $\pm$ 0.45 \\
		& \textbf{TULER-G} & 46.35 $\pm$ 0.60 & 62.61 $\pm$ 0.51 & 30.97 $\pm$ 0.73 & 31.19 $\pm$ 0.84 & 34.33 $\pm$ 0.66 & 39.07 $\pm$ 0.50 & 54.73 $\pm$ 0.37 & 24.78 $\pm$ 0.36 & 25.12 $\pm$ 0.49 & 27.83 $\pm$ 0.41 \\
		& \textbf{Bi-TULER} & 46.93 $\pm$ 0.45 & 63.36 $\pm$ 0.62 & 32.91 $\pm$ 0.54 & 33.42 $\pm$ 0.63 & 35.75 $\pm$ 0.43 & 39.16 $\pm$ 0.54 & 55.46 $\pm$ 0.55 & 26.50 $\pm$ 0.41 & 27.38 $\pm$ 0.59 & 29.17 $\pm$ 0.39 \\
		& \textbf{AttnTUL} & 40.23 $\pm$ 0.50 & 52.15 $\pm$ 0.84 & 25.16 $\pm$ 0.58 & 25.22 $\pm$ 0.78 & 27.84 $\pm$ 0.67 & 41.35 $\pm$ 0.57 & 53.05 $\pm$ 0.43 & 26.90 $\pm$ 0.66 & 27.30 $\pm$ 0.86 & 28.93 $\pm$ 0.50 \\
		& \textbf{CACSR} & 37.11 $\pm$ 1.03 & 47.58 $\pm$ 0.82 & 24.81 $\pm$ 1.06 & 25.44 $\pm$ 1.10 & 26.56 $\pm$ 0.95 & 36.43 $\pm$ 0.93 & 46.15 $\pm$ 0.71 & 23.77 $\pm$ 0.83 & 24.71 $\pm$ 0.92 & 25.24 $\pm$ 0.81 \\
		& \textbf{DeepTUL} & 51.76 $\pm$ 0.61 & 62.61 $\pm$ 0.86 & 27.18 $\pm$ 0.55 & 25.86 $\pm$ 0.58 & 28.65 $\pm$ 0.57 & 46.99 $\pm$ 0.50 & \underline{59.65 $\pm$ 0.40} & 26.39 $\pm$ 0.25 & 25.04 $\pm$ 0.28 & 27.90 $\pm$ 0.31 \\
		& \textbf{S2TUL-HRS} & 42.27 $\pm$ 0.17 & 58.39 $\pm$ 0.10 & 28.15 $\pm$ 0.16 & 28.35 $\pm$ 0.21 & 31.05 $\pm$ 0.22 & 37.73 $\pm$ 0.04 & 50.20 $\pm$ 0.13 & 23.45 $\pm$ 0.03 & 23.37 $\pm$ 0.07 & 26.36 $\pm$ 0.05 \\
		& \textbf{S2TUL-R} & \underline{52.90 $\pm$ 0.45} & \underline{64.73 $\pm$ 0.47} & \underline{39.42 $\pm$ 0.46} & \underline{39.71 $\pm$ 0.65} & \underline{41.81 $\pm$ 0.44} & \underline{48.15 $\pm$ 0.38} & 59.54 $\pm$ 0.39 & \underline{34.65 $\pm$ 0.38} & \underline{34.86 $\pm$ 0.39} & \underline{37.09 $\pm$ 0.41} \\
		& \textbf{HGTUL} & \textbf{54.26 $\pm$ 0.45} & \textbf{66.22 $\pm$ 0.38} & \textbf{41.66 $\pm$ 0.35} & \textbf{41.93 $\pm$ 0.41} & \textbf{45.04 $\pm$ 0.51} & \textbf{51.18 $\pm$ 0.37} & \textbf{62.92 $\pm$ 0.33} & \textbf{36.86 $\pm$ 0.41} & \textbf{37.07 $\pm$ 0.48} & \textbf{39.64 $\pm$ 0.39} \\
		& \textbf{Improvement} & 2.57 & 2.83 & 5.68 & 5.59 & 7.73 & 6.29 & 5.48 & 6.37 & 6.34 & 6.88 \\
		\midrule
		\multirow{9}{*}{\textbf{NYC}} & \textbf{TULER-L} & $36.12 \pm 0.59$ & $46.19 \pm 0.40$ & $20.69 \pm 0.51$ & $22.61 \pm 0.68$ & $22.12 \pm 0.42$ & 30.71 $\pm$ 0.31 & 40.39 $\pm$ 0.19 & 16.93 $\pm$ 0.18 & 18.55 $\pm$ 0.31 & 18.25 $\pm$ 0.14 \\
		& \textbf{TULER-G} & $35.34 \pm 0.61$ & $45.21 \pm 0.51$ & $20.00 \pm 0.65$ & $21.90 \pm 0.96$ & $21.36 \pm 0.56$ & $30.37 \pm 0.60$ & $39.97 \pm 0.36$ & $16.60 \pm 0.37$ & $18.14 \pm 0.37$ & $17.97 \pm 0.47$ \\
		& \textbf{Bi-TULER} & $36.01 \pm 0.48$ & $46.70 \pm 0.63$ & $20.88 \pm 0.34$ & $22.91 \pm 0.65$ & $21.80 \pm 0.31$ & $30.42 \pm 0.63$ & $40.41 \pm 0.68$ & $17.29 \pm 0.43$ & $19.64 \pm 0.53$ & $18.33 \pm 0.42$ \\
		& \textbf{AttnTUL} & $47.02 \pm 0.58$ & $51.63 \pm 0.46$ & $29.83 \pm 1.36$ & $31.79 \pm 2.11$ & $31.43 \pm 0.91$ & $40.05 \pm 0.73$ & $46.50 \pm 0.55$ & $24.16 \pm 1.42$ & $25.79 \pm 2.02$ & $25.71 \pm 0.95$ \\
		& \textbf{CACSR} & 35.95 $\pm$ 0.92 & 45.55 $\pm$ 0.81 & 21.39 $\pm$ 0.35 & 22.78 $\pm$ 0.78 & 22.32 $\pm$ 0.68 & 34.18 $\pm$ 0.58 & 42.36 $\pm$ 0.48 & 20.45 $\pm$ 0.45 & 21.71 $\pm$ 0.48 & 21.41 $\pm$ 0.50 \\
		& \textbf{DeepTUL} & \underline{51.05 $\pm$ 0.94} & \underline{57.83 $\pm$ 0.71} & $27.50 \pm 1.04$ & $26.44 \pm 1.19$ & $28.65 \pm 0.95$ & \underline{44.07 $\pm$ 0.48} & \underline{52.86 $\pm$ 0.71} & $24.97 \pm 0.37$ & $23.98 \pm 0.42$ & $26.06 \pm 0.38$ \\
		& \textbf{S2TUL-HRS} & 44.72 $\pm$ 0.49 & 53.31 $\pm$ 0.27 & 28.04 $\pm$ 0.37 & 28.84 $\pm$ 0.53 & 30.85 $\pm$ 0.41 & 35.91 $\pm$ 0.65 & 45.56 $\pm$ 0.22 & 20.11 $\pm$ 0.56 & 20.82 $\pm$ 0.78 & 23.32 $\pm$ 0.58 \\
		& \textbf{S2TUL-R} & $48.95 \pm 0.43$ & $56.15 \pm 0.49$ & \underline{33.13 $\pm$ 0.65} & \underline{34.62 $\pm$ 0.78} & \underline{34.88 $\pm$ 0.67} & $42.70 \pm 0.30$ & $51.36 \pm 0.49$ & \underline{27.72 $\pm$ 0.42} & \underline{28.70 $\pm$ 0.50} & \underline{29.32 $\pm$ 0.39} \\
		& \textbf{HGTUL} & \textbf{54.26 $\pm$ 0.22} & \textbf{61.16 $\pm$ 0.20} & \textbf{38.41 $\pm$ 0.40} & \textbf{39.00 $\pm$ 0.50} & \textbf{41.38 $\pm$ 0.26} & \textbf{48.11 $\pm$ 0.25} & \textbf{57.64 $\pm$ 0.21} & \textbf{32.49 $\pm$ 0.34} & \textbf{33.39 $\pm$ 0.45} & \textbf{35.07 $\pm$ 0.29} \\
		& \textbf{Improvement} & $6.29$ & $5.76$ & $15.94$ & $12.65$ & $18.64$ & 9.17 & 9.04 & 17.21 & 16.34 & 19.61 \\
		\midrule
		\multirow{9}{*}{\textbf{JKT}} & \textbf{TULER-L} & $34.73 \pm 0.57$ & $50.03 \pm 0.70$ & $22.43 \pm 0.71$ & $24.62 \pm 1.04$ & $24.67 \pm 0.52$ & 25.94 $\pm$ 0.67 & 37.93 $\pm$ 0.46 & 16.81 $\pm$ 0.43 & 19.10 $\pm$ 0.55 & 18.40 $\pm$ 0.44 \\
		& \textbf{TULER-G} & 33.97 $\pm$ 0.74 & 49.49 $\pm$ 0.65 & 21.25 $\pm$ 0.62 & 23.16 $\pm$ 0.72 & 23.66 $\pm$ 0.64 & 26.10 $\pm$ 0.45 & 38.22 $\pm$ 0.67 & 16.64 $\pm$ 0.19 & 18.74 $\pm$ 0.43 & 18.36 $\pm$ 0.26 \\
		& \textbf{Bi-TULER} & 35.83 $\pm$ 0.51 & 50.95 $\pm$ 0.56 & 24.15 $\pm$ 0.80 & 26.62 $\pm$ 0.93 & 26.20 $\pm$ 0.75 & 26.91 $\pm$ 0.60 & 39.25 $\pm$ 0.50 & 18.29 $\pm$ 0.41 & 21.32 $\pm$ 0.52 & 19.58 $\pm$ 0.46 \\
		& \textbf{AttnTUL} & 31.62 $\pm$ 1.16 & 45.15 $\pm$ 0.81 & 20.53 $\pm$ 1.01 & 22.28 $\pm$ 1.11 & 21.98 $\pm$ 1.02 & 33.37 $\pm$ 0.75 & 45.96 $\pm$ 0.79 & 23.25 $\pm$ 0.71 & 25.15 $\pm$ 0.73 & 24.50 $\pm$ 0.67 \\
		& \textbf{CACSR} & 34.56 $\pm$ 0.94 & 46.15 $\pm$ 0.98 & 25.01 $\pm$ 1.07 & 27.82 $\pm$ 1.17 & 25.64 $\pm$ 1.11 & 34.86 $\pm$ 1.35 & 44.30 $\pm$ 0.65 & 24.56 $\pm$ 1.42 & 26.60 $\pm$ 1.33 & 25.62 $\pm$ 1.47 \\
		& \textbf{DeepTUL} & \underline{47.62 $\pm$ 0.97} & \underline{59.42 $\pm$ 0.86} & 31.65 $\pm$ 0.95 & 30.92 $\pm$ 0.91 & 32.42 $\pm$ 1.11 & \underline{44.68 $\pm$ 0.43} & \underline{53.88 $\pm$ 0.39} & 29.43 $\pm$ 0.46 & 28.70 $\pm$ 0.55 & 30.19 $\pm$ 0.38 \\
		& \textbf{S2TUL-HRS} & 32.95 $\pm$ 0.67 & 46.15 $\pm$ 0.16 & 24.80 $\pm$ 0.76 & 29.74 $\pm$ 1.04 & 26.11 $\pm$ 0.80 & 26.83 $\pm$ 0.15 & 37.40 $\pm$ 0.11 & 18.52 $\pm$ 0.09 & 21.47 $\pm$ 0.28 & 20.29 $\pm$ 0.15 \\
		& \textbf{S2TUL-R} & 46.87 $\pm$ 0.35 & 58.52 $\pm$ 0.66 & \underline{38.23 $\pm$ 0.58} & \underline{41.75 $\pm$ 0.75} & \underline{38.83 $\pm$ 0.44} & 40.65 $\pm$ 0.34 & 51.21 $\pm$ 0.40 & \underline{31.27 $\pm$ 0.42} & \underline{34.09 $\pm$ 0.65} & \underline{31.92 $\pm$ 0.35} \\
		& \textbf{HGTUL} & \textbf{57.19 $\pm$ 0.47} & \textbf{66.34 $\pm$ 0.57} & \textbf{48.17 $\pm$ 0.50} & \textbf{50.97 $\pm$ 0.52} & \textbf{49.98 $\pm$ 0.51} & \textbf{48.95 $\pm$ 0.21} & \textbf{59.83 $\pm$ 0.21} & \textbf{37.80 $\pm$ 0.25} & \textbf{40.50 $\pm$ 0.30} & \textbf{39.83 $\pm$ 0.24} \\
		& \textbf{Improvement} & $20.09$ & $11.64$ & $26.00$ & $22.08$ & $28.71$ & 9.56 & 11.04 & 20.88 & 18.80 & 24.78 \\
		\bottomrule
	\end{tabular}
\end{adjustbox}
\end{table*}

\textbf{Q1.} We conduct a comprehensive performance comparison between our proposed model and baseline models across three datasets, with the results summarized in \autoref{tab:performance}. The best results are highlighted in \textbf{bold}, while the second-best results are underlined. The experimental results demonstrate that our model significantly outperforms all baseline models across all evaluation metrics. Specifically, compared to the second-best baseline models, our model achieves improvements in ACC@1 ranging from 2.57\% to 20.09\%, and in Macro-F1, the improvements range from 5.68\% to 26.00\%.

Further analysis of the three variants of the \textbf{TULER} model— \textbf{TULER-L}, \textbf{TULER-G}, and \textbf{Bi-TULER} provides additional insights. The results reveal that the LSTM-based variants, \textbf{TULER-L} and \textbf{Bi-TULER}, consistently outperform the GRU-based \textbf{TULER-G} across all datasets. This observation underscores the stronger capacity of LSTM models in capturing complex trajectory patterns, which is beneficial to addressing the challenges inherent in the TUL.

Among the baseline models, \textbf{DeepTUL} and \textbf{S2TUL-R} stand out, achieving the second-best results. \textbf{DeepTUL} is an attentive recurrent neural network. Although traditional RNN architectures struggle to capture global information between trajectories, \textbf{DeepTUL} introduces a historical attentive module to capture the multi-periodic patterns of user movement from labeled historical trajectories. It partially leverages relational features between trajectories for learning. By using the spatio-temporal features of POIs in historical trajectories as input, \textbf{DeepTUL} highlights the essential role of spatio-temporal information and trajectory movement patterns in the TUL task.
\textbf{S2TUL-R}, on the other hand, models the repeatability relationships between trajectories by constructing a repeatability graph and employs a GCN for learning. This approach directly captures global relational between trajectories, achieving excellent performance. These results further underscore the importance of global relational information in addressing the challenges of the TUL problem.

Despite its strong performance in \textbf{ACC@k}, \textbf{DeepTUL} lags behind \textbf{S2TUL-R} in \textbf{Macro-F1}, likely due to the imbalanced distribution of user data in the TUL task. This imbalance prevents DeepTUL from effectively learning the features of minority-class users, thereby impacting the \textbf{Macro-F1} score. The effect of data balancing on \textbf{DeepTUL} performance will be further discussion in \autoref{sec:databal}.

Moreover, comparing our model to other baselines, we observe an average improvement of 15.34\% in \textbf{Macro-F1} and 8.99\% in \textbf{ACC@1}. The significant improvement in \textbf{Macro-F1} indicates that our model effectively balance the prediction performance across different users, especially in handling rare users, details will be discussed in \autoref{sec:coldstart}. This highlights the robustness and generalization ability of our model in handling imbalanced data within the TUL task.
%Moreover, the average improvement in \textbf{Macro-R} is 17.72\%, surpasses that of \textbf{Macro-F1}, it also indicates that the model performs well on cold-start users linking. This highlights the robustness and generalization ability of our model in handling imbalanced data within the TUL task.
Finally, we can observe that the performance of all models decreases as the number of users increases. This is because an increase in the number of users complicates the classification task, thereby reducing linking accuracy. However, our model maintains optimal performance even with an increasing number of users, often exhibiting the smallest decline in linking accuracy, further demonstrating the stability of \textbf{HGTUL} performance with respect to the number of users.

%\begin{table}[hb]
%	\centering
%	\caption{Average Trajectory Numbers for Different User Types}
%	\label{tab:avelen}
%	\begin{tabular}{@{}lcccccc@{}}
%		\toprule
%		\multirow{2}{*}{\textbf{User type}} & \multicolumn{2}{c}{\textbf{GOWALLA}} & \multicolumn{2}{c}{\textbf{NYC}} & \multicolumn{2}{c}{\textbf{JKT}} \\
%		\cmidrule(lr){2-3} \cmidrule(lr){4-5} \cmidrule(l){6-7}
%		&500 & 1000 &500 & 1000 & 500 &1000 \\
%		\midrule
%		Active & 10 & 11 & 15 & 15 & 16 & 15 \\
%		Normal & 4 & 4 & 6 & 6 & 7 & 7 \\
%		Inactive & 2 & 2 & 3 & 3 & 4 & 4 \\
%		\bottomrule
%	\end{tabular}
%\end{table}
\vspace{-1em}
\subsection{Ablation Study}

\begin{table*}[t]
	\centering
	\caption{Results of Ablation Study}
	\label{tab:ablation_study}
	\begin{adjustbox}{width=\textwidth}
		\begin{tabular}{@{}lcc|cc|cc@{}}
			\toprule
			& \multicolumn{2}{c}{\textbf{Gowalla} (Users=1000)} & \multicolumn{2}{c}{\textbf{NYC} (Users=1000)} & \multicolumn{2}{c}{\textbf{JKT} (Users=500)} \\
			\cmidrule(lr){2-3} \cmidrule(lr){4-5} \cmidrule(lr){6-7}
			\textbf{Methods} & \textbf{ACC@1 (\%)} & \textbf{Macro-F1 (\%)} & \textbf{ACC@1 (\%)} & \textbf{Macro-F1 (\%)} & \textbf{ACC@1 (\%)} & \textbf{Macro-F1 (\%)} \\
			\midrule
			\textbf{HGTUL-A}  & 50.05 $\pm$ 0.22 & 35.61 $\pm$ 0.42 & 46.74 $\pm$ 0.20 & 30.78 $\pm$ 0.19 & 55.88 $\pm$ 0.17 & 46.93 $\pm$ 0.15 \\
			\textbf{HGTUL-Ap} & 50.96 $\pm$ 0.26 & 37.05 $\pm$ 0.47 & 47.74 $\pm$ 0.22 & 32.40 $\pm$ 0.29 & 56.67 $\pm$ 0.20 & 47.94 $\pm$ 0.32 \\
			\textbf{HGTUL-S}  & 47.28 $\pm$ 0.48 & 34.06 $\pm$ 0.43 & 38.60 $\pm$ 0.53 & 23.94 $\pm$ 0.51 & 46.31 $\pm$ 0.42 & 37.96 $\pm$ 0.53 \\
			\textbf{HGTUL-L}  & 48.90 $\pm$ 0.32 & 34.36 $\pm$ 0.58 & 48.04 $\pm$ 0.23 & 32.36 $\pm$ 0.20 & 55.78 $\pm$ 0.39 & 45.98 $\pm$ 0.64 \\
			\textbf{HGTUL-H}  & 41.87 $\pm$ 0.40 & 25.86 $\pm$ 0.39 & 31.85 $\pm$ 0.43 & 14.75 $\pm$ 0.39 & 39.88 $\pm$ 0.53 & 28.06 $\pm$ 0.56 \\
			\textbf{HGTUL-D}  & 47.85 $\pm$ 0.15 & 27.51 $\pm$ 0.11 & 45.32 $\pm$ 0.22 & 24.01 $\pm$ 0.25 & 55.26 $\pm$ 0.53 & 42.82 $\pm$ 0.72 \\
			\midrule
			\textbf{HGTUL}    & {51.18 $\pm$ 0.37} & {36.86 $\pm$ 0.41} & {48.11 $\pm$ 0.25} & {32.49 $\pm$ 0.34} & {57.19 $\pm$ 0.47} & {48.17 $\pm$ 0.50} \\
			\bottomrule
		\end{tabular}
	\end{adjustbox}
\end{table*}

\textbf{Q2.} To evaluate the effectiveness of each module in \textbf{HGTUL}, we conduct an ablation study, with the results presented in \autoref{tab:ablation_study}. First, to assess the role of multi-perspective trajectory representations, we remove the attentive trajectory representation \textbf{HGTUL-A}, the structural trajectory representation \textbf{HGTUL-S}, and the spatio-temporal trajectory representation \textbf{HGTUL-L}. The results indicate that removing any of these representations causes a performance decline, with the structural trajectory representation having the greatest impact. The contributions of the spatio-temporal and attentive representations vary across datasets. Specifically, spatio-temporal features are important for the TUL in the \textbf{GOWALLA} (1000users) and \textbf{JKT} (500users) datasets, while meaningful POI features play a more important role in the \textbf{NYC} (1000users) dataset, reflecting the differences in dataset characteristics.

To further validate the effectiveness of our attentive trajectory representation learning method, we conduct an additional ablation study \textbf{HGTUL-Ap}, where the attentive trajectory representation is obtained by aggregating vertex features based on attention scores. The results show that our approach, which directly learns trajectory embeddings through dynamic updates, outperforms \textbf{HGTUL-Ap}. However, the performance of \textbf{HGTUL-Ap} still shows improvement compared to \textbf{HGTUL-A}, demonstrating the significance of attentive trajectory representations in the model.

Next, to explore the contribution of hypergraph modeling, we remove the trajectory hypergraph module \textbf{HGTUL-H}. The results demonstrate a significant performance decline, highlighting the necessity of modeling complex high-order inter-trajectory association relationships for the TUL task. Finally, to examine the effect of data balancing, we remove the data balancing module \textbf{HGTUL-D}. The results show performance degradation across all datasets, with a notable drop in \textbf{Macro-F1} for the \textbf{NYC} (1,000 users) and \textbf{GOWALLA} (1,000 users) datasets, demonstrating the importance of data balancing for model robustness, particularly in larger user populations.

Overall, among all modules, the trajectory hypergraph exhibits the most significant impact when removed, further confirming the effectiveness of hypergraph-based trajectory modeling.

\subsection{User Cold-Start Analysis}
\label{sec:coldstart}
\begin{figure}
	\centering
	\includegraphics[width=\linewidth]{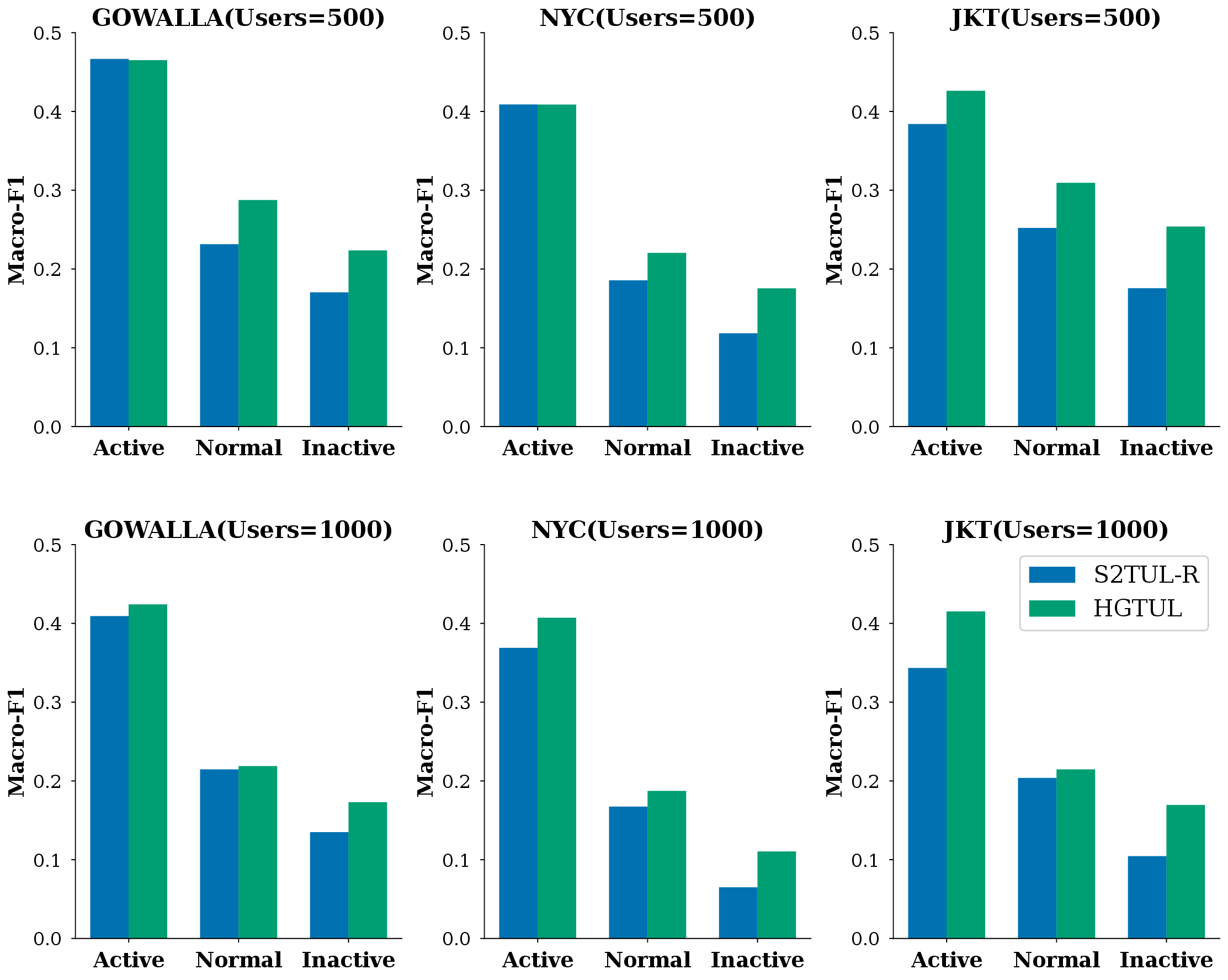}
	\caption{User Cold-Start Performance Comparison: HGTUL vs. S2TUL-R}
	\label{fig:coldstart}
\end{figure}
\textbf{Q3.} To further assess our model's ability to alleviate the user cold-start problem, we consider the sparsity issue inherent in check-in data, where the cold-start problem is particularly pronounced in TUL tasks based on such data. It is especially challenging to accurately identify the user associated with the trajectory for new users who lack historical trajectory information. We categorize users into three groups: active, normal, and inactive, with classification based on the number of user trajectories. In the training set, the top 30\% of users, the number of trajectories, are classified as active users, the bottom 30\% as inactive users, and the remaining as normal users. We use the best-performing baseline model, \textbf{S2TUL-R}, for comparison. \autoref{fig:coldstart} illustrates the comparison of cold-start performance, with \textbf{Macro-F1} score as the evaluation metric. Experimental results show that the \textbf{HGTUL} outperforms \textbf{S2TUL} in most cases, with the most significant improvement observed for inactive users. This indicates that our model can effectively address the cold-start problem by capturing high-order inter-trajectory relationships. Furthermore, on the NYC and JKT datasets, as the number of users increases, the model's performance improvement becomes more pronounced, suggesting that with more trajectories, more inter-trajectory relationships can be learned in the hypergraph, thereby improving the linking accuracy.

\subsection{Data Balancing Analysis}
\label{sec:databal}
\begin{figure}
	\centering
	\includegraphics[width=\linewidth]{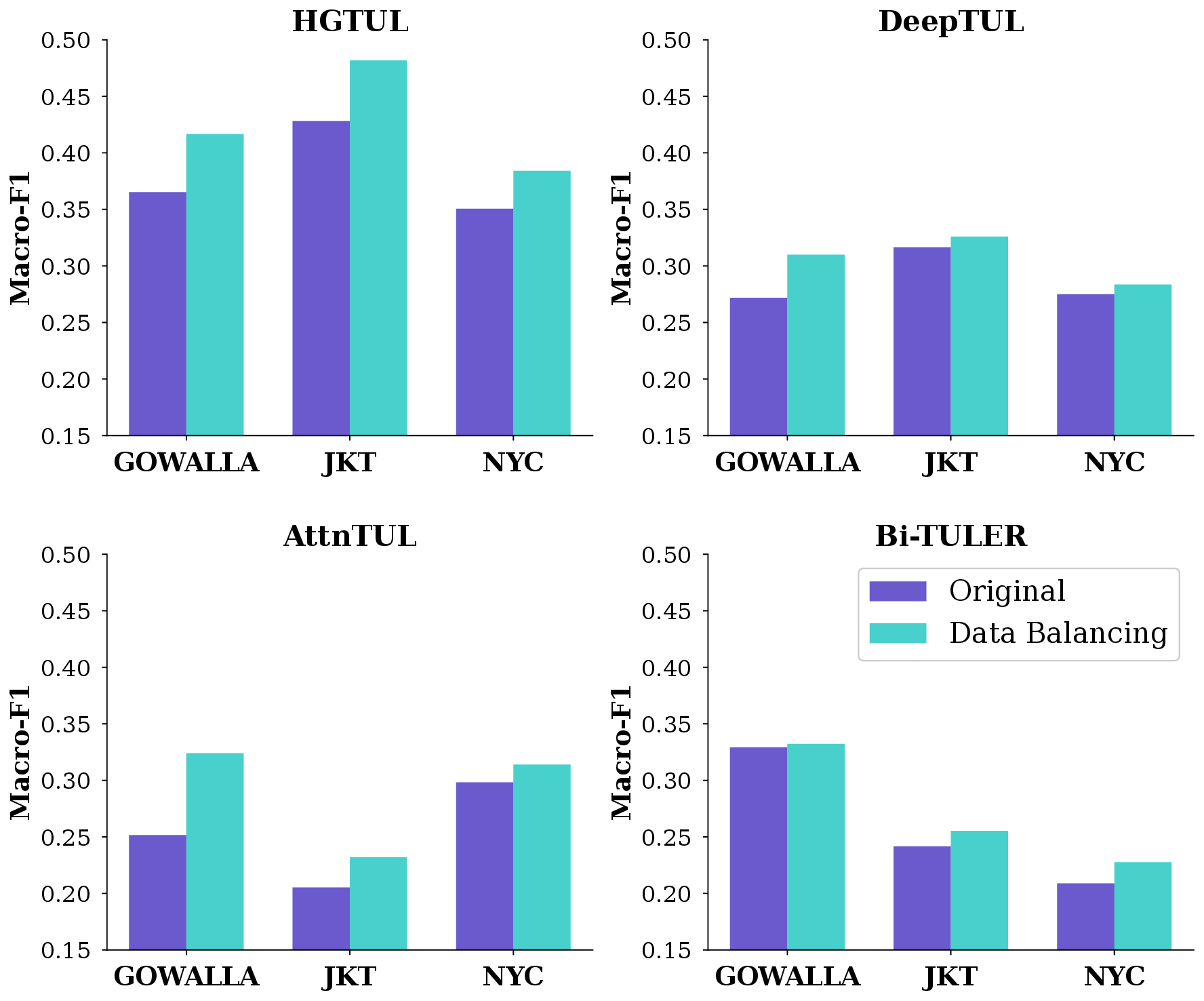}
	\caption{Impact of Data Balancing on Algorithm Performance.}
	\label{fig:databal}
	\vspace{-2em}
\end{figure}

\textbf{Q4.} To investigate the impact of data balancing on the performance of the TUL model, we apply the proposed data balancing method to \textbf{HGTUL} and three baseline models. Since \textbf{Macro-F1} provides a comprehensive measure of multi-class classification performance, we adopt it as the evaluation metric. First, as shown in \autoref{fig:databal}, data balancing enhances the TUL classification performance across all models and datasets, highlighting its essential role in TUL model training—an aspect often overlooked in previous studies. Second, among the baseline models, \textbf{AttnTUL} achieves the most significant improvement through data balancing. By analyzing its architecture, we find that \textbf{AttnTUL} employs a \textbf{GNN} to learn POI representations. Under imbalanced data distributions, the model tends to associate high-frequency user check-in behaviors with specific POIs, leading to incorrect trajectory-user linking for minority users. This makes \textbf{AttnTUL} more sensitive to data imbalance. Finally, after applying data balancing, \textbf{HGTUL} still outperforms other baseline models, demonstrating that our model achieves superior classification performance by effectively integrating high-order association relationships and spatio-temporal information.

\section{Conclusion}
In this work, we propose a novel hypergraph-based multi-perspective trajectory user linking model to address the limitations of existing methods in capturing high-order relationships among trajectories and the variable influence of POIs. By constructing a trajectory hypergraph and leveraging a hypergraph attention network, our model effectively learns the high-order inter-trajectory associations and variable impact of POIs on trajectories from relational perspective. And we incorporate the temporal and spatial information of trajectories as sequence features into an LSTM network from spatio-temporal perspective. Additionally, the proposed data balancing method significantly improves the model's performance on inactive users and demonstrates the necessity of addressing class imbalance in TUL tasks. Extensive experiments on three real-world datasets show that HGTUL outperforms state-of-the-art baselines, achieving remarkable improvements in ACC@1 and Macro-F1 metrics. 
Our work presents a method for TUL and suggests the potential of modeling high-order relationships in trajectory data analysis. Future work may investigate extending the proposed framework to other spatio-temporal data mining tasks.
\balance

%\begin{acks}
% This work was supported by the [...] Research Fund of [...] (Number [...]). Additional funding was provided by [...] and [...]. We also thank [...] for contributing [...].
%\end{acks}

%\clearpage

\bibliographystyle{ACM-Reference-Format}
\bibliography{sample}

\end{document}